\documentclass[11pt,a4paper]{article}
\usepackage[hyperref]{emnlp2020-templates/emnlp2020}
\usepackage{times}
\usepackage{latexsym}

\usepackage{microtype}

\usepackage{booktabs}
\usepackage{amsmath}

\aclfinalcopy 

\newcommand{\up}[1]{${\uparrow}#1$}
\newcommand{\down}[1]{${\downarrow}#1$}
\newcommand{\eg}[0]{e.g.}                

\usepackage{multirow}
\usepackage{rotating}
\usepackage{enumitem}
\usepackage{tipa}

\usepackage{xcolor}

\usepackage[T2A,T1]{fontenc}
\usepackage[utf8]{inputenc}
\usepackage[russian,english]{babel}
\usepackage{paralist}

\title{It's not a Non-Issue: \\ Negation as a Source of Error in Machine Translation}

\author{Md Mosharaf Hossain\mbox{\normalfont,}\textsuperscript{\textipa{8}}
        Antonios Anastasopoulos\mbox{\normalfont,}\textsuperscript{\textipa{Z}}
        Eduardo Blanco\mbox{\normalfont,}\textsuperscript{\textipa{8}} \and
        Alexis Palmer\textsuperscript{\textipa{P}}\\
\textsuperscript{\textipa{8}}Department of Computer Science and Engineering, University of North Texas\\
\textsuperscript{\textipa{Z}}Department of Computer Science, George Mason University\\
\textsuperscript{\textipa{P}}Department of Linguistics, University of North Texas\\
{\footnotesize
\texttt{mdmosharafhossain@my.unt.edu} \hspace{.2cm}
\texttt{\{eduardo.blanco,alexis.palmer\}@unt.edu} \hspace{.2cm}
\texttt{antonis@gmu.edu}}}

\date{}

\begin{document}
\maketitle
\begin{abstract}
As machine translation (MT) systems progress at a rapid pace, questions of their adequacy linger. In this study we focus on negation, a universal, core property of human language that significantly affects the semantics of an utterance. 
We investigate whether translating negation is an issue for modern MT systems using 17 translation directions as test bed. Through thorough analysis, we find that indeed the presence of negation can significantly impact downstream quality, in some cases resulting in quality reductions of more than 60\%.
We also provide a linguistically motivated analysis that directly explains the majority of our findings.
We release our annotations and code to replicate our analysis here: \url{https://github.com/mosharafhossain/negation-mt}.
\end{abstract}

\section{Introduction}\label{sec:intro}

Machine translation (MT) has come a long way, steadily improving the quality of automatic systems, relying mostly on the advent of neural techniques~\cite{goldberg2017neural} and availability of large data collections.
As neural MT (NMT) systems start to become ubiquitous, however, one should take note of whether the improvements, as measured by evaluation campaigns such as WMT~\cite[\textit{et alia}]{ws-2017-machine,ws-2018-machine-translation-shared,ws-2019-machine-translation} or IWSLT shared tasks~\cite[\textit{et alia}]{cettolo2017overview,niehues2018iwslt} or other established benchmarks, do in fact mean better quality, especially with regards to the semantic adequacy of translations.

We focus specifically on negation, a single yet quite complex phenomenon, which can severely impact the semantic content of an utterance. As NMT systems increasingly improve, to the point of claiming human parity in some high-resource language pairs and in-domain settings \cite{hassan2018achieving},\footnote{We direct the reader to \cite{laubli-etal-2018-machine} and \cite{toral-etal-2018-attaining} for further examination of such claims.} fine-grained semantic differences become increasingly important.
Negation in particular, with its property of logical reversal, has the potential to cause loss of (or mis-)information if mistranslated.

Other linguistic phenomena and analysis axes have gathered significant attention in NMT evaluation studies, including anaphora resolution~\cite{hardmeier2014translating,voita-etal-2018-context} and pronoun translation~\cite{guillou-hardmeier-2016-protest}, modality~\cite{baker2012modality}, ellipsis and deixis~\cite{voita-etal-2019-good}, word sense disambiguation~\cite{tang-etal-2018-analysis}, and morphological competence~\cite{burlot-yvon-2017-evaluating}. Nevertheless, the last comprehensive study of the effect of negation in MT pertains to older, phrase-based models~\cite{fancellu-webber-2015-translating-negation}.
In this work, we set out to study the effect of negation in modern NMT systems. Specifically, we explore: 
\begin{compactenum}
    \item Whether negation affects the quality of the produced translations~(it does); 
    \item Whether the typically-used evaluation datasets include a significant amount of negated examples~(they don't);
    \item Whether different systems quantifiably handle negation differently across different settings~(they do); and
    \item Whether there is a linguistics-grounded explanation of our findings~(there is).
\end{compactenum}

Our conclusion is that indeed negation still poses an issue for NMT systems in several language pairs, an issue which should be tackled in future NMT systems development. Negation should be taken into consideration especially when deploying real-world systems which might produce incredibly fluent but inadequate output~\cite{martindale-etal-2019-identifying}.

\section{Negation 101}\label{sec:negation}

Negation at its most straightforward---simple negation of declarative sentences---involves reversing the truth value of a sentence.
\emph{Skies are blue}, for example, becomes \emph{Skies are not blue}.
\textbf{Clausal negation} of an existing utterance, defined roughly as negation of the entire clause rather than a subpart, produces a second utterance whose meaning contradicts that of the first.\footnote{See \citet{penka2015negation} and \citet{horn2010expression}, among others, for more on the semantics, syntax, and interpretation of negation.}
Without stipulations about special contexts (e.g. referring to \emph{skies} in two different places), it is not possible for both utterances to be true. 
\textbf{Constituent negation} involves negation of a non-clausal constituent, such as \emph{happy} in \emph{The universe is not happy with us}.
Finally, \textbf{lexical negation} refers to two different phenomena.
The first are words formed by adding negative affixes, such as \emph{non-starter} or \emph{unhappy}.
The second are words with no negative affixes whose meanings nonetheless convey negation, such as the verb \emph{lack} in \emph{This engine lacks power}.

Negation is a core property of human language; every language provides mechanisms for expressing negation, and, cross-linguistically, there are a small number of strategies used for its expression \cite{dahl1979typology,miestamo2007negation,horn2010expression}.
Following \citet{wals-112}, languages may use three different mechanisms for clausal negation: a) \textbf{negative particles} (e.g. English \emph{not} or German \emph{nicht}); b) \textbf{negative morphemes or affixes} (e.g. Lithuanian \emph{ne-}); and/or c) \textbf{negative auxiliary verbs}, (e.g. Finnish \emph{ei}), a strategy in which (generally) inflection moves from the main verb to the auxiliary verb.   
Depending on the language, constituent and lexical negation may be coded using the same mechanisms as clausal negation, or with entirely different mechanisms.
Negative pronouns (e.g. English \emph{nothing} or \emph{nowhere}), negative adverbials (\emph{never}), and negative polarity items (\emph{any}) add additional complexity to the expression of negation.

Even within negation, cross-linguistic variability is immense, ranging from different linguistic strategies, to different structural relationships between linguistic elements, even to differences in interpretation. 
\citet{szabolcsi2004conjunction} show that the interpretation of conjunction, disjunction, and negation can differ cross-linguistically. 
In English ``not X and Y'' can be interpreted as either 
$\neg X\wedge\neg Y$ (``neither X nor Y'') or $\neg X \vee \neg Y$ (``not X or not Y'') while in Hungarian (and Russian, Italian, etc.) only the first interpretation is allowed. 
Such differences are potential error sources for translating negation.

\paragraph{Negation-related translation errors.}
Here we
show examples of different types of errors that 
NMT systems make when translating negation.
    
    \textbf{1. Negation Omission}: The system output completely omits the translation of the negation cue. 
    The following reference translation from the Turkish-English WMT18 shared task reads: ``[...] Don't run for public office, if you can't take heat from voters.'' The best performing system, though, outputs ``[...] if you can't take criticism from voters, you're a candidate for state duty,'' contradicting the reference translation.\footnote{The source sentence in Turkish is: ``Eğer seçmenlerin eleştirilerini kaldıramıyorsan devlet görevine aday olma.''}
    
    \textbf{2. Negation Reversal}: The semantic meaning is reversed, so that the sentence ends up meaning the opposite of the intended meaning, often in fine-grained ways. Consider the following reference translation from WMT19 Lithuanian-English: ``The family lawyer of the deceased biker also asks for reversal of the verdict of not guilty.'' Here the ``verdict of not guilty'' entails that there was no conviction. The output of the translation system, however, implies there was a conviction to be reversed: ``The family lawyer of the dead rider also asks for the conviction to be lifted.''\footnote{The Lithuanian source is: ``Panaikinti išteisinam\k{a}j\k{i} nuosprend\k{i} prašo ir žuvusio motociklininko šeimos advokatė.''}
    
    \textbf{3. Incorrect Negation Scope:} The system output makes errors in argument mapping, such that the wrong constituent is negated. Here we look at an example from Finnish-English WMT18. Differences between the reference translation (``The reason is not the Last Judgment'') and the best performing system output (``The last judgment is not the reason'') lead to differences in interpretation.\footnote{Finnish source sentence: ``Viimeinen tuomio ei ole syy.''}
    
    \textbf{4. Mistranslation of Negated Object:} When the negated element in the sentence is wrongly translated, meaning is disturbed. This example comes from WMT18 German-English. The reference translation begins ``No exchange of personal data occurred [...],'' and the system output for the best system begins ``There was no exchange of personnel [...].''\footnote{The source sentence in German is: ``Zu einem Personalienaustausch kam es aber nicht, da der 75-Jährige die Dame auf dem Parkplatz nicht mehr finden konnte.''} Aside from the mistranslated object noun phrase, the meaning is intact.
\noindent

These error types vary in their severity, but each has the potential to completely change the meaning conveyed by the translation.

\section{Experimental Setup}\label{sec:setup}
We follow the setup of the Conference on Machine Translation~(WMT),
in particular, the 2018 and 2019 Shared Tasks~\cite{bojar-etal-2018-findings,barrault-etal-2019-findings}.
We compare reference and system translations using normalized direct assessments.
Direct assessments are scores from 0 to 100 provided by researchers or crowd workers.
In order to account for the hundreds of human annotators with potentially different criteria,
raw direct assessments are normalized using the mean and standard deviation of each annotator.
The normalized direct assessment is the average of the sentence-level direct assessments.
We refer to normalized direct assessment with Z-score or simply~Z.

Z-scores are the official ranking criterion in WMT competitions,
and are preferred to automated metrics to assess the quality of MT.
Nevertheless, most of the MT community still relies on automated metrics for development and system comparisons.
Thus, we also work with three automated metrics,
in particular
BLEU~\cite{post-2018-call},
chrF++~\cite{popovic-2017-chrf},
and METEOR~\cite{denkowski-lavie-2011-meteor}.

In the remainder of the paper we present
two complementary analyses.
First, we investigate 
the role of negation in machine translation
with an emphasis on numeric evaluation (Section~\ref{sec:quant}).
Second, we investigate from a 
linguistic perspective what makes translating negation difficult (Section~\ref{sec:ling}).

\noindent
\textbf{Datasets.}
We work with all submissions to the news translation tasks in the WMT18 and WMT19 competitions.
Table~\ref{t:ranking_all} shows the language directions in each competition along with the number of sentences with Z-scores in the corresponding test set.
Specifically, we investigate systems for both translation directions between English and Russian (\textsc{ru}), Estonian (\textsc{et}), German (\textsc{de}), Turkish (\textsc{tr}), Finnish (\textsc{fi}), Czech (\textsc{cs}), and Chinese (\textsc{zh}). We also use Lithuanian (\textsc{lt}), Gujarati (\textsc{gu}), and Kazakh (\textsc{kk}) to English systems from WMT 19.

\noindent
\textbf{Negation Detection.}
Due to the lack of reliable negation detection systems in most languages, our study is limited to focusing on cases where negation is present in English.
This creates slightly different settings for translation into and out of English, hence we distinguish them in our analyses and presentation of the results.
For translation out of English, we detect negation cues in the source sentence.
For translation into English, we 
detect negation cues in the reference translations. 

In order to detect negation automatically,
we train a negation cue detector for English 
using a Bi-LSTM neural network with a CRF layer, as described in~\cite{hossain-etal-2020-predicting}.
Trained and evaluated with a publicly available corpus~\cite{morante-blanco-2012-sem},
it obtains 0.92 F1.
The cue detector recognizes single-token cues
(\emph{not}, \emph{n't}, \emph{never}, \emph{no}, \emph{nothing}, \emph{nobody}, etc.)
as well as affixal cues (\emph{im}possible, \emph{dis}agree, fear\emph{less}, etc.).
The supplemental materials provide details about the architecture and input representation of the cue detector. 

\paragraph{Important note.}
We make the strong assumption in our analyses that presence of a negation cue in the English reference translation indicates presence of negation in the source sentence.
We acknowledge that investigating the role of negation in machine translation
by only looking at English negations likely misses valid insights. 
For example, 
a Spanish sentence containing negation
(e.g., \emph{``El ladrón no estaba preocupado hasta que vino la policía''})
can be translated into English either with negation (\textit{``The thief was not worried until the police arrived''}), or without negation
(\emph{``The thief only worried when the police arrived''}).
We reserve for future work a more thorough analysis of correspondences between negation in source sentences and negation in English reference translations.

\section{Quantitative Analysis}\label{sec:quant}
We conduct a thorough quantitative analysis of the WMT18 and WMT19 submissions
around six questions investigating the role of negation in MT.

\begin{table}[ht!]
\small
\centering

\begin{tabular}{r@{$\rightarrow$}l rr@{\ }rr@{\ }r}
\toprule
\multicolumn{7}{c}{Translation \textit{into} English}\\
\multicolumn{2}{c}{}
  & \multicolumn{1}{c}{all} &
    \multicolumn{2}{c}{w/ neg.} &
    \multicolumn{2}{c}{w/o neg.} \\ \cmidrule(lr){3-3} \cmidrule(lr){4-5} \cmidrule(lr){6-7}
\multicolumn{2}{c}{WMT18} &
  \multicolumn{1}{c}{Z} & \multicolumn{1}{c}{Z} & \multicolumn{1}{c}{\%$\Delta$} & \multicolumn{1}{c}{Z} & \multicolumn{1}{c}{\%$\Delta$} \\
\midrule

ru & en &  0.215 & 0.132 & ($-$38.6) & 0.230 & ($+$7.0)\\
de & en &  0.413 & 0.376 & ($-$9.0) & 0.421 & ($+$1.9) \\ 
et & en &  0.326 & 0.232 & ($-$28.8) & 0.343 & ($+$5.2)\\ 

tr & en &  0.045 & 0.014 & ($-$68.9) & 0.051 & ($+$13.3)\\

fi & en &  0.153 & 0.223 & ($+$45.8) & 0.139 & ($-$9.2)\\

cs & en &  0.298 & 0.333 & ($+$11.7) & 0.290 & ($-$2.7)\\ 

zh & en &  0.140 & 0.162 & ($+$15.7) & 0.137 & ($-$2.1)\\  
\midrule
\multicolumn{2}{c}{WMT19} \\
ru & en & 0.156 & 0.123 & ($-$21.2) & 0.161 & ($+$3.2) \\
de & en & 0.146 & 0.136 & ($-$6.9) & 0.147 & ($+$0.7) \\
fi & en & 0.285 & 0.306 & ($+$7.4) & 0.283 & ($-$0.7) \\
lt & en & 0.234 & 0.093 & ($-$60.3) & 0.262 & ($+$12.0) \\
gu & en & 0.210 & 0.112 & ($-$46.7) & 0.221 & ($+$5.2) \\
kk & en & 0.270 & 0.326 & ($+$20.7) & 0.264 & ($-$2.2) \\

\end{tabular}
\begin{tabular}{r@{$\rightarrow$}l rr@{\ }rr@{\ }r}
\toprule
\multicolumn{7}{c}{Translation \textit{from} English}\\ 
\multicolumn{2}{c}{}
  & \multicolumn{1}{c}{all} &
    \multicolumn{2}{c}{w/ neg.} &
    \multicolumn{2}{c}{w/o neg.} \\ 
    \cmidrule(lr){3-3} \cmidrule(lr){4-5} \cmidrule(lr){6-7} 
\multicolumn{2}{c}{WMT18} &
  \multicolumn{1}{c}{Z} & \multicolumn{1}{c}{Z} & \multicolumn{1}{c}{\%$\Delta$} & \multicolumn{1}{c}{Z} & \multicolumn{1}{c}{\%$\Delta$} \\
\midrule
en & ru &  0.352 & 0.245 & ($-$30.4) & 0.371 & ($+$5.4) \\ 
en & de &  0.653 & 0.689 & ($+$5.5) & 0.646 & ($-$1.1) \\ 
en & et &  0.549 & 0.439 & ($-$20.0) & 0.569 & ($+$3.6) \\
en & tr &  0.277 & 0.094 & ($-$66.1) & 0.308 & ($+$11.2) \\ 
en & fi &  0.521 & 0.569 & ($+$9.2) & 0.512 & ($-$1.7) \\ 
en & cs &  0.594 & 0.574 & ($-$3.4) & 0.599 & ($+$0.8) \\ 
en & zh &  0.219 & 0.229 & ($+$4.6) & 0.218 & ($-$0.5) \\ 
\bottomrule
\end{tabular}

\caption{Evaluation of the best WMT18 and WMT19 submissions (normalized direct assessments, Z) for each language direction using all sentences, sentences with negation (w/ neg.), and sentences without negation (w/o neg.).
Translating sentences with negation obtains worse results in many languages
(e.g., Turkish, Russian, Estonian, Lithuanian, Gujarati).
}
\label{t:best_all}
\vspace{-1em}
\end{table}

\paragraph{Q0: Is Negation Present in Evaluation Datasets?}
We found that 9.6--20\% of source sentences contain negation depending on the target language.
These percentages
are lower than what we observe in
online reviews~(23--29\%),
books~(29\%)
and conversation transcripts~(27--30\%) \cite{hossain-etal-2020-analysis},
and are also lower than previous reports \cite{morante-sporleder-2012-modality}.

\paragraph{Q1: Is Translating Sentences with and without Negation Equally Hard?}
Table \ref{t:best_all} shows the Z-scores obtained by the best submission in each translation direction
using all sentences, sentences with negation, and sentences without negation.
Many language pairs obtain substantially worse Z-scores for sentences containing negation:
Turkish, Russian, and Estonian in WMT18 (20.0--68.9\% lower),
and Lithuanian, Gujarati, and Russian in WMT19 (21.2--60.3\% lower).
When both translation directions are available for a language pair (WMT18),
translating negation into English from these languages consistently receives lower scores than translating from English.

Interestingly, 
sentences with negation receive better Z-scores in Finnish, Chinese, and Kazakh (4.6--45.8\% higher).
Finally, only two languages show opposite trends translating from and into English negations: German and Czech, although the differences in Z-scores are smaller (-3.4--11.7\%).

Naturally, other factors beyond the presence of negation can affect the results. Since sentence length tends to negatively correlate with translation quality, the length difference between sentences with and without negation could be an important factor. Sentences with negation are on average longer than sentences without negation,\footnote{This is expected if one considers the additional tokens required, such as the negation cue, or other potential syntactic changes (e.g. adding an auxiliary verb).} but we do not consider the differences (typically within 2-6 words) to be significant. Nevertheless, we replicated our analysis from Table~\ref{t:best_all} using only the sentences that fall within a standard deviation of the mean sentence length $[\mu\pm\sigma]$ for each dataset. The results, shown in Table~\ref{t:eval_bucket_1} in the Appendix, do not differ significantly from our Table~\ref{t:best_all} analysis. We identify one major inconsistency, in the case of English-Finnish translation direction; we attribute it to issues in identifying negation in that dataset, and we leave further analysis as future work. In any case, performing the same analysis on sentences that fall outside that bucket (that is, sentences shorter or longer than one standard deviation of the average $(0,\mu-\sigma)$ and $(\mu+\sigma,+\infty)$, respectively, available in Tables~\ref{t:eval_bucket_3} and~\ref{t:eval_bucket_2} in the Appendix) does not yield conclusive results. We attribute this to the fact that these buckets include fewer data samples and too many outliers (of very easy or very hard sentences). 

In any case, the consistency of the results in the ``average'' case leads us to conclude: a) that \textbf{negation affects Z-scores in all language pairs and directions}, and b) \textbf{that translating from or into English sentences containing negation in most language directions is harder than sentences without negation.}

\begin{table}[t]
\small
\centering
\begin{tabular}{r@{$\rightarrow$}l@{\ \ }rccc}
\toprule
\multicolumn{6}{c}{Translation \textit{into} English} \\ 
\multicolumn{3}{l}{WMT18} & w/ neg. & w/o neg. & w/ v. w/o \\
\midrule
ru & en & (1,820) & 0.786 & 1.000 & 0.786\\
de & en & (2,121) & 0.933 & 0.983 & 0.917\\ 
et & en & (1,593) & 0.729 & 0.972 & 0.692 \\ 
tr & en & (2,678) & \underline{0.738} & \underline{0.800}  & \underline{0.527}\\                      
fi & en & (1,769) & 0.833 & 0.944 & 0.778\\ 
cs & en & (1,849) & 1.000 & 1.000 & 1.000\\ 
zh & en & (2,039) & 0.495 & 0.906 & 0.398 \\
\midrule
\multicolumn{3}{l}{WMT19} \\
ru & en & (1,692) & 0.575 & 0.912 & 0.486\\
de & en & (2,000) & 0.723 & 0.941 & 0.650\\
fi & en & (1,548) & 0.879 & 0.970 & 0.848\\
lt & en & (1,000) & 0.855 & 0.871 & 0.722\\ 
gu & en & (1,016) & 0.745 & 1.000 & 0.745\\
kk & en & (1,000) & 0.855 & 1.000 & 0.855\\ \toprule
\multicolumn{6}{c}{Translation \textit{from} English} \\ 
\multicolumn{3}{l}{WMT18} & w/ neg. & w/o neg. & w/ v. w/o \\
\midrule
en & ru & (2,076) & 0.944 & 1.000 & 0.944 \\
en & de & (759) & 0.917 & 0.983  & 0.900 \\
en & et & (1,019) & 0.978 & 1.000 & 0.978 \\
en & tr & (420) & 0.643 & 0.929  & \underline{0.571} \\
en & fi & (760) & 0.848 & 0.970  & 0.818 \\
en & cs & (1,594) & 1.000 & 1.000 & 1.000 \\
en & zh & (1,876) & 0.758 & 0.956 & 0.714 \\ 
\bottomrule
\end{tabular}

\caption{Comparison of rankings (Kendall's $\tau$) of all submissions to WMT18 and WMT19 using
  (a) all sentences and those with negation (w/ neg.),
  (b) all sentences and those without negation (w/o neg.),
  and
  (c) sentences with and without negation (w/ v. w/o).
  The numbers between parentheses indicate the number of sentences with human scores for the best system.
  All $\tau$ coefficients are statistically significant ($p < 0.05$)
  except those that are underlined (null hypothesis: there is no association between the rankings).
  We note, however, that the $\tau$ coefficients are substantially lower in most language directions when negation is present.
}
\label{t:ranking_all}
\end{table}

\begin{table*}[ht!]
\small
\centering
\begin{tabular}{l@{\ }rlr
                r clr
                r clr}
\toprule

& 
\multicolumn{3}{c}{All sentences} &
\multicolumn{4}{c}{Sentences w/ negation} &
\multicolumn{4}{c}{Sentences w/o negation} \\ \cmidrule(lr){2-4} \cmidrule(lr){5-8} \cmidrule(lr){9-12}
& & \multicolumn{1}{c}{System} & \multicolumn{1}{c}{Z} &
  & & \multicolumn{1}{c}{System} & \multicolumn{1}{c}{Z} &
  & & \multicolumn{1}{c}{System} & \multicolumn{1}{c}{Z} \\

\midrule

\multirow{8}{*}{\begin{sideways}WMT18\end{sideways}}
& 1 & Alibaba & 0.215  & 1 &  \up{1} & online-B & 0.172 & 1 & -- & Alibaba & 0.230 \\ 

& 2 & online-B & 0.192 & 2 &  \up{1} & online-G & 0.155 & 2 & -- & online-B & 0.196 \\ 

& 3 & online-G & 0.170 & 3 &     \down{2} & Alibaba & 0.132 & 3 & -- & online-G & 0.173 \\ 

& 4 & uedin & 0.110 &   4 & -- & uedin & 0.084 & 4 & -- & uedin & 0.115 \\ 

& 5  & online-A & 0.034 &  5 & -- & online-A & 0.009 & 5 & -- & online-A & 0.039 \\ 

& 6  & afrl & -0.014 &  6 & \up{1} & JHU & -0.044 &  6 & -- & afrl & 0.012 \\ 

& 7  & JHU & -0.027 &  7 & \down{1} & afrl & -0.154 & 7 & -- & JHU & -0.024 \\ 

\midrule

\multirow{14}{*}{\begin{sideways}WMT19\end{sideways}}
& 1  & FB-FAIR & 0.156 & 
  1 & -- & FB-FAIR & 0.123 & 
  1 & -- & FB-FAIR & 0.161 \\ 

& 2 & online-G & 0.134 & 
  2 & \up{2} & online-B & 0.091 &  
  2 & -- & online-G & 0.143 \\ 

& 3 & eTranslation & 0.122 & 
  3 & \up{6} & online-A & 0.077 & 
  3 & -- & eTranslation & 0.134 \\ 

& 4 & online-B & 0.121 &  
  4 & \down{2} & online-G & 0.073 & 
  4 & \up{2} & MSRA.SCA & 0.133 \\ 

& 5 & NEU & 0.115 &   
  5 & \down{2} & eTranslation & 0.044 &   
  5 & -- & NEU & 0.126 \\ 

& 6 & MSRA.SCA & 0.102 &   
  6 & \down{1} & NEU & 0.040 &   
  6 & \down{2} & online-B & 0.125 \\ 

& 7 & rerank-re & 0.084 &   
  7 & \up{1} & online-Y & -0.023 &   
  7 & -- & rerank-re & 0.108 \\ 

& 8 & online-Y & 0.076 &   
  8 & \up{4} & TartuNLP-u & -0.046 &   
  8 & -- & online-Y & 0.091 \\ 

& 9 & online-A & 0.029 &   
  9 & \up{1} & afrl-sys & -0.066 &   
  9 & \up{1} & afrl-sys & 0.024 \\ 

& 10 & afrl-sys & 0.012 &   
  10 & \down{3} & rerank-re & -0.072 &   
  10 & \down{1} & online-A & 0.022 \\ 

& 11 & afrl-ewc & -0.039 &   
  11 & -- & afrl-ewc & -0.101 &
  11 & -- & afrl-ewc & -0.030 \\

& 12 &  TartuNLP-u & -0.040 &   
  12 & \up{1} & online-X & -0.101 &   
  12 & -- & TartuNLP-u & -0.039 \\ 

& 13 & online-X & -0.097 &   
  13 & \down{7} & MSRA.SCA & -0.102 &   
  13 & -- & online-X.0 & -0.096 \\ 


\bottomrule

\end{tabular}

\caption{Rankings of all submissions translating from Russian to English using
all sentences (official WMT ranking),
sentences with negation,
and sentences without negation.
FB-FAIR, afrl, and afrl-sys refer to Facebook-FAIR, afrl-ruen-syscomb, and afrl-syscomb19.
We use \up{k} (\down{k}) to indicate the gains (losses) in absolute ranking with respect to the ranking obtained  with all sentences.
For example, in WMT19, MSRA.SCA is 7 positions lower in the ranking obtained with sentences with negation (from 6th to 13th),
and online.A is 6 positions higher (from 9th to 3rd).
As the $\tau$ coefficients indicate (Table \ref{t:ranking_all}),
there are barely any changes in the ranking obtained with sentences without negation, but many changes in the ranking obtained with sentences with negation.
}
\label{t:ranking_one_pair}
\end{table*}

\paragraph{Q2: Does Negation Affect Rankings?}
Just because the best system in each language direction obtains better or worse Z-scores (Q1),
it is not necessarily the case that \emph{all} systems do better or worse.
In order to check if rankings are affected,
we calculate the correlation between rankings
obtained with Z-scores and
(a)~sentences with negation and
(b)~sentences without negation (outlined in Table~\ref{t:ranking_all}).
We use Kendall's $\tau$ coefficient \cite{kendall1938new},
which only considers the ranking---not the differences in Z-scores.
$\tau$ coefficients range from~-1 to~1 (absolute negative and positive correlation),
and $\tau=0$ indicates no correlation at all.

We observe that the Z-score rankings for all sentences and sentences without negation are
very close in all language directions: $\tau\!\in\!(0.8, 1.0)$, and all but two
(\mbox{tr$\rightarrow$en} and \mbox{lt$\rightarrow$en}) are above~$0.9$.
The rankings obtained based on sentences with negation, on the other hand, show lower $\tau$ correlations, except language pairs involving German and Czech.
Except those involving German and Chinese, all language pairs have at least one translation direction
with $\tau\!\le\!0.85$.
We also observe that the rankings change much more (lower $\tau$ coefficients) translating 
from Chinese, Estonian, and Russian
into English sentences containing negation 
than translating from English sentences containing negation into those languages.

It is worth noting that the relative drop in Z-score of the best system when negation is present (\%$\Delta$, Table \ref{t:best_all})
is not a good predictor of ranking changes.
For example,
all submissions translating from English to Russian obtain proportionally worse Z-scores when negation is present in the source text, thus the ranking barely changes ($\tau\!=\!0.944$) despite $\%\Delta\!=\!-30.4$.
On the other hand,
we observe many ranking changes translating from Chinese to English ($\tau\!=\!0.495$)
and vice versa ($\tau\!=\!0.758$), despite $\%\Delta$ of only~15.7 and~4.6.

In Table \ref{t:ranking_one_pair}, we show the ranking of submissions
translating from Russian to English
obtained with all sentences (official WMT ranking),
and the rankings obtained with sentences containing and not containing negation.
The ranking changes in WMT18 (\up{} and \down{} arrows) illustrate the $\tau$ correlation coefficients
obtained with sentences
with negation (many changes, $\tau\!=\!0.575$)
and
without negation (few changes, $\tau\!=\!0.912$). 
The supplementary materials
contain similar tables for selected language pairs in WMT18 and WMT 19.

We conclude that \textbf{rankings based on sentences containing negation are substantially different for most translation directions. Thus, different systems behave differently translating from and into English sentences containing negation.}

\begin{figure}[t]
\centering
\includegraphics[width=.9\columnwidth]{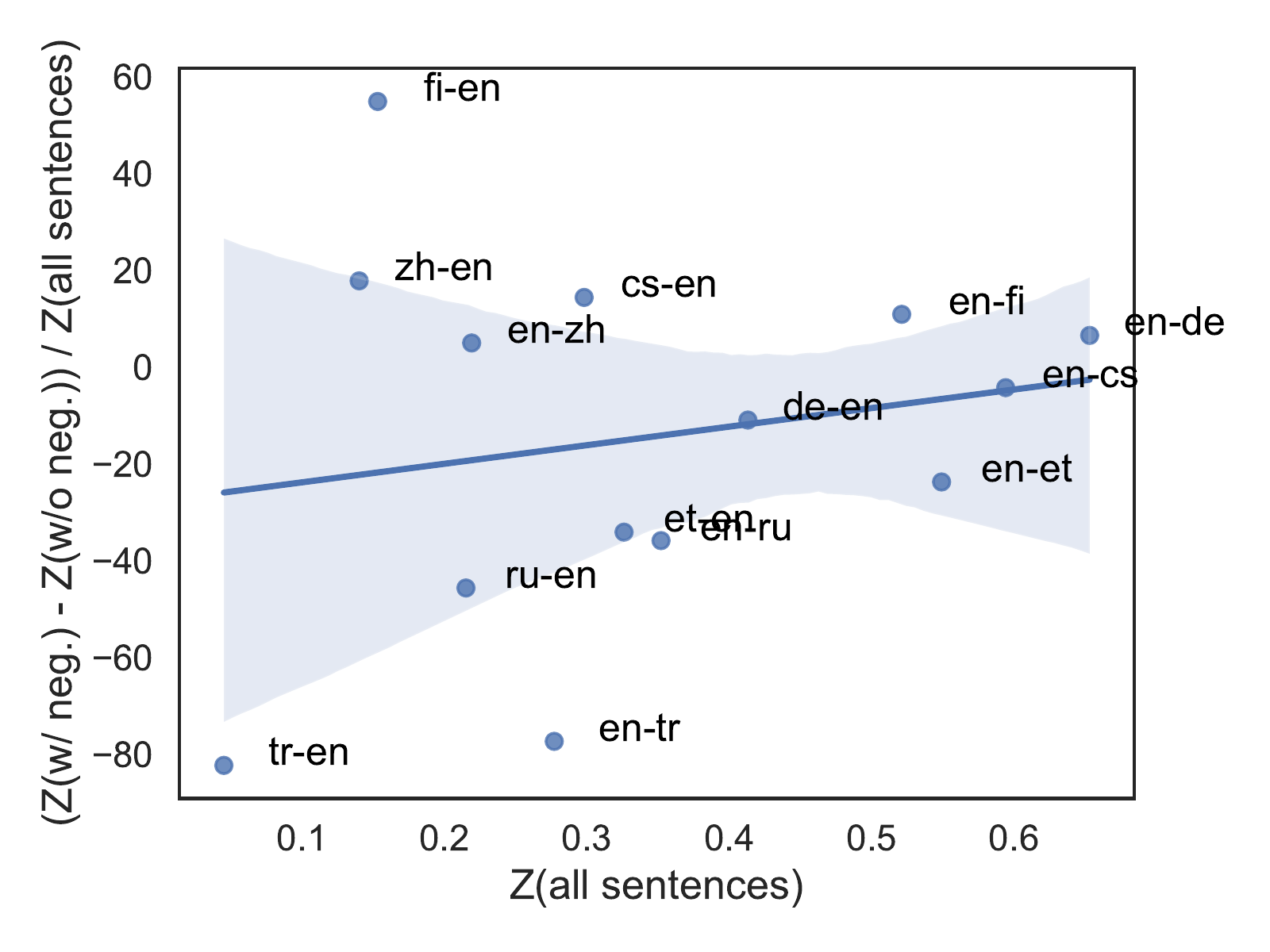}
\vspace{-1em}
\caption{Correlation between the Z-score with all sentences and the \textit{relative} drop for the sentences with negation (best system per language pair).
}
\label{f:all_diff}
\vspace{-1em}
\end{figure}

\paragraph{Q3: Is Translating Negation Harder with Harder Language Pairs and Directions?}
Not all translation directions are equally easy to model,
as evidenced by the wide variance of direct assessment scores in the WMT competitions (Table~\ref{t:best_all}).
Figure~\ref{f:all_diff} shows that there is is a weak positive correlation between the \emph{relative} differences in Z-score with negation and the overall Z-score.
There are however many exceptions/outliers, e.g.,
translations from Finnish or Russian into English receive roughly the same Z-scores,
but negation is much harder from Russian than from Finnish.
We conclude that \textbf{the difficulty of translating between two languages is only a weak indicator
of how difficult it is to translate negation, most notable with overall Z-scores below 0.3.}

\begin{figure}[t]
\centering
\includegraphics[width=.9\columnwidth]{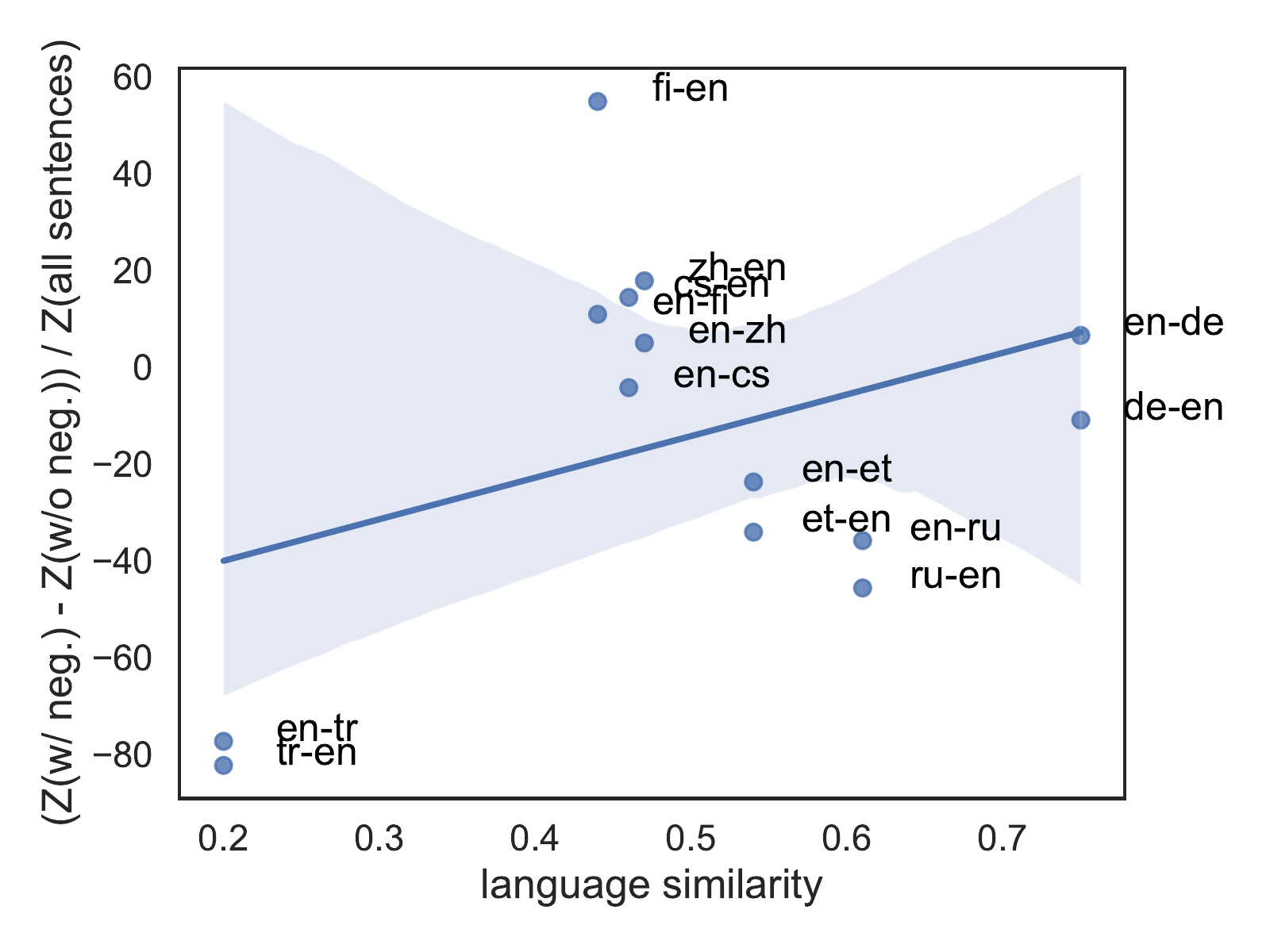}
\vspace{-1em}
\caption{Correlation between the language similarity and the \textit{relative} drop for the sentences with negation.}
\label{f:similarity_diff}
\vspace{-1em}
\end{figure}

\paragraph{Q4: Is Translating Negation between Similar Languages Easier?}
Intuition may lead us to believe that it is easier to translate negation between similar languages.
We show the correlation between language similarity and relative differences in Z-scores with and without negation
in Figure~\ref{f:similarity_diff}.
To calculate similarity between two languages, we follow \newcite{zhang-toral-2019-effect} and \newcite{berzak2017predicting}.
Briefly, we obtain feature vectors for each language from lang2vec~\cite{Littel-et-al:2017},
and define the similarity between two languages as the cosine similarity between their feature vectors.
More specifically, we concatenate
103 morphosyntactic features and 87 language family features (only those relevant to the languages we work with) from the URIEL typological database. 

We conclude that \textbf{similarity between languages is only a weak indicator of how difficult it is to translate negation}.
We revisit this question in Section~\ref{sec:ling} with an in-depth linguistic discussion.

\begin{table*}[ht!]
\small
\centering

\setlength{\tabcolsep}{.077in}
\small
\begin{tabular}{r@{$\rightarrow$}lrrrrrr c r@{$\rightarrow$}lrrrrrr}
\toprule
\multicolumn{2}{c}{} & \multicolumn{6}{c}{Z-score v.} &
\multicolumn{3}{c}{} & \multicolumn{6}{c}{Z-score v.} \\ \cmidrule(lr){3-8}  \cmidrule(lr){12-17}
\multicolumn{2}{c}{} & \multicolumn{2}{c}{BLEU} &
                       \multicolumn{2}{c}{chrF++} &
                       \multicolumn{2}{c}{METEOR} &
                       &
\multicolumn{2}{c}{} & \multicolumn{2}{c}{BLEU} &
                       \multicolumn{2}{c}{chrF++} &
                       \multicolumn{2}{c}{METEOR} \\ 
                       \cmidrule(lr){3-4} \cmidrule(lr){5-6} \cmidrule(lr){7-8}
                       \cmidrule(lr){12-13} \cmidrule(lr){14-15} \cmidrule(lr){16-17}

\multicolumn{2}{c}{}  & \multicolumn{1}{c}{w/} & \multicolumn{1}{c}{w/o} & \multicolumn{1}{c}{w/} & \multicolumn{1}{c}{w/o} &  \multicolumn{1}{c}{w/} & \multicolumn{1}{c}{w/o}  &
&
\multicolumn{2}{c}{}  & \multicolumn{1}{c}{w/} & \multicolumn{1}{c}{w/o} & \multicolumn{1}{c}{w/} & \multicolumn{1}{c}{w/o} &  \multicolumn{1}{c}{w/} & \multicolumn{1}{c}{w/o} \\

\midrule

ru & en & 0.64 & 0.79 & 0.78 & 0.86 & 0.79 & 0.79 &&
en & ru & 0.83 & 0.94 & 0.83 & 0.94 & 0.83 & 0.94 \\

de & en & 0.80 & 0.85 & 0.83 & 0.92 & 0.78 & 0.88 &&
en & de & 0.78 & 0.75 & 0.78 & 0.77 & 0.75 & 0.77 \\

et & en & 0.78 & 0.83 & 0.71 & 0.85 & 0.71 & 0.85 &&
en & et & 0.76 & 0.85 & 0.89 & 0.89 & 0.82 & 0.93 \\

tr & en & \underline{-0.32} & \underline{-0.60} & \underline{-0.74} & \underline{-0.20} & \underline{-0.53} & \underline{0.00} &&
en & tr & 0.79 & \underline{0.57} & 0.64 & 0.86 & 0.64 & 0.79 \\

fi & en & 0.78 & 0.78 & 0.78 & 0.94 & 0.72 & 0.94 &&
en & fi & 0.85 & 0.82 & 0.67 & 0.94 & 0.79 & 0.91 \\

cs & en & \underline{0.60} & \underline{0.80} & \underline{0.80} & \underline{0.80} & 1.00 & \underline{0.80} &&
en & cs & 1.00 & 1.00 & 1.00 & 1.00 & 1.00 & 1.00 \\

zh & en & 0.47 & 0.80 & 0.52 & 0.73 & 0.47 & 0.82 &&
en & zh & 0.41 & \underline{0.39} & 0.69 & 0.65 & \underline{-0.16} & \underline{0.32} \\ \bottomrule

\end{tabular}

\caption{
  Comparison of WMT18 rankings (Kendall's $\tau$) using several metrics, and sentences with and without negation.
  We compare the official scoring (normalized direct assessments, Z-score)
  and three automated metrics: BLEU, chrF++ and METEOR.
  We provide the same analysis for WMT19 in the supplementary materials.
  All $\tau$ coefficients are statistically significant ($p<0.05$) except those that are underlined
  (null hypothesis: there is no association between the Z-score and the automatic metric).
  The differences in the rankings obtained with the Z-score and the three metrics are more substantial when negation is present (lower $\tau$ coefficients).
  }
\label{t:da_bleu_meteor_wmt18}
\end{table*}

\paragraph{Q5: Are Automatic Metrics Worse with Negation?}
Machine translation evaluation is an active field of research,
and new automatic metrics are proposed yearly \cite[\textit{et alia}]{fomicheva2019taking}.
The ideal metric would correlate perfectly with human judgments,
and increased correlation is often a justification for new metrics \cite{lavie-agarwal-2007-meteor}.
We investigate whether three popular metrics (BLEU, chrF++ and METEOR)
are equally suitable when applied to sentences with and without negation.
Note that the smallest change when negation is present (e.g., dropping \emph{never} or \emph{n't})
is likely to result in a (very) bad translation.
Table~\ref{t:da_bleu_meteor_wmt18} shows the correlations  (Kendall's $\tau$) between the rankings on WMT 18, obtained
with the Z-scores and the three metrics using
(a)~sentences with negation,
and
(b)~sentences without negation.

When negation is present,
the three metrics obtain worse correlation coefficients or just slightly better (within 5\%).
This is true with all translation directions except those involving Finnish or Turkish.
The drops in correlation coefficients are substantial with the three metrics in many translation directions, e.g.,
from Russian, Chinese or Gujarati into English.
While there is no winner across all language directions
(e.g., chrF++ is better with Russian, and METEOR with Czech),
the correlation coefficients show, unsurprisingly,
that BLEU is the least suited to evaluate translation quality when negation is present.

We conclude that \textbf{automatic metrics are bad estimators of machine translation quality when negation is present, and that chrF++ and METEOR are better suited than BLEU.}
\section{So what's going wrong?}\label{sec:ling}

\begin{table*}[t]
\small
\centering



\begin{tabular}{l|cccc|c||l|cr} \toprule
 & \multicolumn{4}{c}{(a) Typological properties} & & \multicolumn{3}{c}{(b) Negation-related translation errors} \\ \midrule
Language & Clausal & PredNeg & Symm & Order & Sim-EN & Direction & neg.sents & \%neg.errors \\ \midrule
Turkish & Affix & Yes & Both & [V-Neg] & 1.0 & tr $\rightarrow$ en     & 460 & 17.2\% \\ 
Lithuanian  & Affix & Yes & -- & [Neg-V] & 1.0 & lt $\rightarrow$ en & 164 & 11.0\%\\ 
Russian & Particle & Yes & Symm & NegV & 2.5 & ru $\rightarrow$ en & 218 & 8.3\% \\ 
German & Particle & No & Symm & NegV/VNeg & 3.0 & de $\rightarrow$ en & 281 & 7.5\% \\ 
Finnish & Aux. & Yes & Asymm & NegV & 1.5 &  fi $\rightarrow$ en & 185 & 2.7\%\\ \midrule 
English &  Particle & No & Both & NegV & & & \\ \bottomrule
\end{tabular}
\caption{(a) Typological properties related to negation (left), and (b) Error analysis of negation-containing test set sentences (right) for five non-English WMT languages. With the exception of Finnish, languages more similar to English with respect to negation show fewer negation-related errors in translation. We discuss Finnish below.}\label{tab:negErrors}
\vspace{-1em}
\end{table*}

We have shown that translations with negation receive lower scores on average than translations without negation, for most languages. 
We have also shown that this is true regardless of a system's overall performance.
To better understand \emph{why} negation is challenging for NMT systems, we look into the properties of negation in individual languages, as well as the particular errors made by the systems. 
For this analysis, we restrict ourselves to 5 language pairs, mostly selecting languages with substantial negation-related performance differences for one or more of the quantitative analyses (Turkish, Lithuanian, Russian, and Finnish). We also consider German, as English-German/German-English is one of the most commonly used translation benchmarks.  

\subsection{Typological perspective on negation}

The World Atlas of Linguistic Structures \cite[WALS][]{wals} is a typological database of the world's languages.
The core of WALS is a partially-filled grid of typological features vs. languages.
For example,
WALS identifies 5 broad categories of features related to the realization of negation.
The left side of table~\ref{tab:negErrors} shows the relevant feature values for our languages.\footnote{We leave out prohibitives~\cite{wals-71}, as this construction is infrequent and does not affect our analysis.}

\textbf{(i.) Clausal: Form of simple clausal negation.}
This feature captures the morphological form of clausal negation for declarative sentences \cite{wals-112} and has three possible values.
\textbf{Particle} languages use a negation word (e.g. \emph{not} in EN or \emph{nicht} in DE), \textbf{Affix} languages use a negative affix (e.g. the prefix \emph{ne-} in LT), and \textbf{Aux} languages use a negative auxiliary verb.
Aux languages are the rarest; only 47 out of 1157 languages with a value for this property in WALS use a negative auxiliary verb.
Finnish is one of these.
In FI the negative verb (underlined) inflects to agree with the subject, as for example: \emph{Linja-auton kuljettajaa \underline{ei} epäillä rikoksesta}, ``The bus driver is not suspected of a crime.'' 
This means that clausal negation in Finnish declarative sentences is always expressed with the same verb, and the form of the verb relies only on the person and number of the subject; other aspects of the sentence do not change the form of the verb.

\textbf{(ii.) PredNeg: Presence of predicate negation with negative indefinites.}
In some languages, negative indefinite pronouns (e.g. \emph{nothing, nowhere}) require an additional particle negating the predicate \cite{wals-115}.
Russian is an example of a language with the predicate negation requirement: \emph{\foreignlanguage{russian}{Сергей Сироткин: Я \underline{ничего не} мог сделать [...]
}}, ``Sergey Sirotkin: There was nothing I could do [...].'' \foreignlanguage{russian}{ничего} translates as \emph{nothing} in English, and \foreignlanguage{russian}{не} is the predicate negation particle.
Most varieties of English do not use a predicate negation particle, though some do, leading to a value of `No/Mixed' for this feature in WALS.
Because our systems all translate from and into a mainstream variety of English, we use `No' for this analysis.

\textbf{(iii.) Symm: Symmetricity of negation.}
\citet{wals-113} characterizes languages as symmetric or asymmetric with respect to simple clausal negation.
In symmetric languages (e.g. DE and RU), negated clauses are the same as their non-negated counterparts, with the exception of the negation word.
In asymmetric languages (e.g. Finnish), presence of the negation word triggers other grammatical changes in the sentence.
For FI, the negative auxiliary is inflected and the main verb changes to an uninflected form.
Some languages (like English) show both behaviors depending on context.

\textbf{(iv.) Order: Order of negation cues and other constituents.}
Various aspects of the ordering of negation cues in the clause are addressed in two sections of WALS \cite{wals-143,wals-144}.
Here we look only at the relative ordering of the negation cue and the verb.
NegV (negation cue before the verb), the ordering seen in EN, RU, and FI, is the most common (525/1325 languages sampled).
LT also shows NegV ordering, with the difference that the negation cue is an affix rather than a particle (indicated by square brackets and hyphen: [Neg-V]).
TR shows the opposite ordering, and German allows both NegV and VNeg.

\paragraph{Similarity to English.} For this analysis, we assign each language a score for its similarity to English with respect to typological properties of negation, based on WALS data.
English is the comparison language because it is the common language in all translation directions.
For each of the four WALS properties, a language scores one point for a feature with the same value as English, and a half point for a partial match. 
For example, Russian gets a half-point for the Symm feature.

\subsection{Are errors due to negation?}
In a next step, we look closely at negation-containing sentences to determine how often systems get the negation wrong.
For these five language pairs, we extract all test-set sentences with negation cues detected in the reference translations; the number of sentences per language pair is shown in Table~\ref{tab:negErrors}.
After sorting sentences according to Z-score, we compare reference translation and system output and annotate (Yes/No) whether the system gets negation wrong, compared to the reference.
In contrast to \citet{fancellu-webber-2015-translating-negation}, who do fine-grained annotation of translation errors related to negation (focusing on deletion/insertion/substitution of cues, focus, and scope), we ask a broader question designed to capture semantic adequacy focused on handling of negation.
We only choose \textbf{Yes} if the system output shows a glaring error in meaning \textit{related to negation} (see Section~\ref{sec:negation} for discussion of some typical error types).

The percentage of sentences that contain negation with negation-related errors (Table~\ref{tab:negErrors}, right side) ranges from only 2.7\% in Finnish up to more than 17\% in Turkish.
The core finding from this analysis is that (except for Finnish) \textbf{the languages with the fewest negation-related errors are most similar to English with respect to the typology of negation.}

Turkish differs from English on three of four WALS features. In addition, clausal negation in Turkish occurs through a negative affix attached to the verb root.
The complexity of Turkish verbal morphology means that: a) the negative morpheme undergoes changes in form depending on its context; and b) the negative morpheme is tucked away in the interior of the verb word, between the stem and both tense-aspect-mood markers and person agreement morphology \cite{emeksiz2010negation}.
In \emph{Ben seni unut\underline{-ma}-di-m,} (``I have not forgotten you'') \cite{emeksiz2010negation}, the negative morpheme appears as \emph{-ma}.
Affixal clausal negation is not unusual from a typological perspective, but the morphological richness of Turkish makes it particularly difficult to recognize negated clauses.

Lithuanian, Russian, and German each differ from English in their values for 1-2 features.
Interestingly, these languages occupy a sort of middle ground for the percentage of negation errors seen.

Finnish seems to be a special case.
Though it differs from English on 2-3 feature values (2.5 in our scoring system), we see a very low proportion of negation-related errors in system output translations.
We attribute this to the negative auxiliary, a way of expressing negation that is easy to identify, even for non-speakers of the language (and presumably also for NMT systems).
175 out of 185 source sentences contain at least one easily-identifiable negation cue: a conjugated form of the negative auxiliary, prefixal negation on adjectives, a negative conjunction ($\sim$ EN \emph{lest, neither}), or a negative preposition ($\sim$ EN \emph{without}).
We hypothesize that the clarity and detectibility of source-side negation cues improve the quality of NMT systems when translating negation.

\paragraph{Other observations.}
In addition to the main finding above, we notice that, for some languages, certain negation cues are either more or less likely to appear in sentences with negation errors.
For example, in German, negation errors are most likely when the cue is \emph{nicht}-V (negation particle modifying the verb); this is also the most frequent cue overall.
Of 46 source sentences containing the negative indefinite article \emph{kein}, only one triggers a negation error.
We have performed this analysis only for languages where we can reliably (manually) identify negation cues on the source side; we hope to extend to more languages in the future.

Figurative expressions, identified either on the source side or judging by the reference translation, often contain negation errors.
We also see recurring problems with the interaction of negation and the translation of certain temporal expressions, and occasional problems with negation errors in the reference translations~\cite{freitag2020bleu}.
Encouragingly, errors of outright contradiction between the reference translation and system output are rare.

\section{Conclusions}\label{sec:concl}

We show, through both quantitative and qualitative analysis, that negation remains problematic for modern NMT systems.
Though negation is ubiquitous and universal in its semantic effect, its realization varies tremendously from language to language.
Typological similarity with respect to negation seems to correspond to better translation of negation, at least for the language pairs we investigate.
Looking forward, we propose to harness linguistic insights about particular languages to better translate negation, and to devise fine-grained evaluation metrics to capture the adequacy of negation-involving translations.

\section*{Acknowledgements}
The authors are grateful to the anonymous reviewers for their constructive and thorough comments. We also thank Graham Neubig for initial discussions and feedback on the initial stages of the paper.
This material is based in part upon work supported by the National Science Foundation under
Grant No.~1845757.
Any opinions, findings, and conclusions or recommendations expressed in this material
are those of the authors and do not necessarily reflect the views of the NSF.

\bibliography{negation_MT}
\bibliographystyle{emnlp2020-templates/acl_natbib}

\appendix
\section{Identifying Negations}
\label{s:identifying_negs}
\begin{figure}
  \centering
  \input{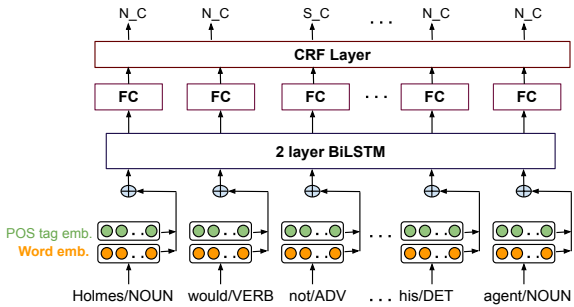}
  \caption{The BiLSTM-CRF model architecture to identify negation cues.
  The input is a sentence, where each token is the actual word and its universal POS tag.
  The model outputs a sequence of labels indication of negation presence.
  The example input sentence is ``Holmes/NOUN would/VERB not/ADV listen/VERB to/ADP such/ADJ fancies/NOUN ,/PUNCT and/CCONJ I/PRON am/VERB his/DET agent/NOUN." }
  \label{fig:nn_architecture}
\end{figure}

In order to identify negations in English sentences
(in source sentences when the translation direction is English to foreign, otherwise in reference translations and system outputs),
we develop a negation cue detector that consists of a two-layer Bidirectional Long Short-Term Memory network with a Conditional Random Field layer on top (BiLSTM-CRF).
This architecture (Figure \ref{fig:nn_architecture}) is similar to the one proposed by \newcite{reimers2017optimal}.
We train and evaluate the model
with CD-SCO, a corpus of Conan Doyle stories with negation annotations~\cite{morante-blanco-2012-sem}.
CD-SCO includes common negation cues (\eg{}, never, no, n't), as well as
prefixal (\eg{}, impossible, unbelievable)
and suffixal negation (\eg{}, motionless).

We map each token in the input sentence to its 300-dimensional pre-trained GloVe embedding \cite{pennington2014glove}. In addition, we extract token-level universal POS tags using spaCy \cite{honnibal2017spacy} and leverage another embedding (300-dimensional) to encode them. Embedding weights for universal POS are learned from scratch as part of the training of the network.
We concatenate the word and POS embeddings, and feed them to the BILSTM-CRF architecture (size of cell state: 200 units).
The representations learnt by the 2-layer BiLSTM are fed to a fully connected layer
with ReLU activation function \cite{nair2010rectified}.
Finally, the CRF layer yields the final output.

We use the following labels to indicate whether a token is a negation cue:
S\_C (single-token negation cue, \eg{}, never, not),
P\_C (prefixal negation, \eg{}, inconsistent),
SF\_C (suffixal negation, \eg{}, emotionless), and
N\_C (not a cue).

\noindent
\textbf{Training details.}
We merge the train and development instances from CD-SCO,
and use 85\% of the result as training and the remaining 15\% as development.
We evaluate our cue detector with the original test split from CD-SCO.
We use the stochastic gradient descent algorithm with 
\mbox{RMSProp} optimizer \citep{tieleman2012lecture} for tuning weights.
We set the batch size to 32,
and the dropout and recurrent dropout are set to 30\% for the LSTM layers.
We stop the training process after the accuracy in the development split does not increase for 20 epochs,
and the final model is the one which yields the highest accuracy in the development accuracy during the training process
(not necessarily the model from the last epoch).
Evaluating with the test set yields the following results:
92.75 Precision, 
92.05. Recall,
and
92.40 F1.
While not perfect, the output of the cue detector is reliable,
and an automatic detector is the only way to count negations in large corpora. The code is available at \url{https://github.com/mosharafhossain/negation-cue}.

Note that our cue detector model has nearly 4.3 million parameters and takes 30 minutes on average to train on a CPU machine (Intel(R) Xeon(R) CPU E5-2680 v4 @ 2.40GHz) with 64 GB of RAM.

\section{Impact of Sentence Length}
\label{s:length_factor}

We present results on the sentence length bucket analysis (discussed in Section~\ref{sec:quant}) in Table \ref{t:eval_bucket_1} (sentences that fall within a standard deviation of the mean sentence length), Table \ref{t:eval_bucket_3} (sentences shorter than one standard deviation of the mean), and Table \ref{t:eval_bucket_2} (sentences longer than one standard deviation of the mean).

\begin{table}[ht!]
\small
\centering
\begin{tabular}{r@{$\rightarrow$}l rr@{\ }rr@{\ }r}
\toprule
\multicolumn{7}{c}{Translation \textit{into} English}\\
\multicolumn{2}{c}{}
  & \multicolumn{1}{c}{all} &
    \multicolumn{2}{c}{w/ neg.} &
    \multicolumn{2}{c}{w/o neg.} \\ \cmidrule(lr){3-3} \cmidrule(lr){4-5} \cmidrule(lr){6-7}
\multicolumn{2}{c}{WMT18} &
  \multicolumn{1}{c}{Z} & \multicolumn{1}{c}{Z} & \multicolumn{1}{c}{\%$\Delta$} & \multicolumn{1}{c}{Z} & \multicolumn{1}{c}{\%$\Delta$} \\
\midrule

ru & en &  0.231 & 0.087 & ($-$62.3) & 0.257 & ($+$11.3)\\
de & en &  0.401 & 0.413 & ($+$3.0)  & 0.399 & ($-$0.5) \\ 
et & en &  0.327 & 0.236 & ($-$27.8) & 0.343 & ($+$4.9)\\ 
tr & en &  0.066 & 0.038 & ($-$42.4) & 0.072 & ($+$9.1)\\
fi & en &  0.149 & 0.172 & ($+$15.4) & 0.144 & ($-$3.4)\\
cs & en &  0.303 & 0.378 & ($+$24.8) & 0.287 & ($-$5.3)\\ 
zh & en &  0.169 & 0.175 & ($+$3.6) & 0.168 & ($-$0.6)\\  
\midrule
\multicolumn{2}{c}{WMT19} \\
ru & en & 0.142 & 0.098 & ($-$31.0) & 0.148 & ($+$4.2) \\
de & en & 0.185 & 0.189 & ($+$2.2) & 0.185 & ($+$0.0) \\

fi & en & 0.294 & 0.272 & ($-$7.5) & 0.297 & ($+$1.0) \\
lt & en & 0.212 & 0.065 & ($-$69.3) & 0.243 & ($+$14.6) \\

gu & en & 0.216 & 0.119 & ($-$44.9) & 0.227 & ($+$5.1) \\
kk & en & 0.286 & 0.384 & ($+$34.3) & 0.276 & ($-$3.5) \\

\end{tabular}
\begin{tabular}{r@{$\rightarrow$}l rr@{\ }rr@{\ }r}
\toprule
\multicolumn{7}{c}{Translation \textit{from} English}\\ 
\multicolumn{2}{c}{}
  & \multicolumn{1}{c}{all} &
    \multicolumn{2}{c}{w/ neg.} &
    \multicolumn{2}{c}{w/o neg.} \\ 
    \cmidrule(lr){3-3} \cmidrule(lr){4-5} \cmidrule(lr){6-7} 
\multicolumn{2}{c}{WMT18} &
  \multicolumn{1}{c}{Z} & \multicolumn{1}{c}{Z} & \multicolumn{1}{c}{\%$\Delta$} & \multicolumn{1}{c}{Z} & \multicolumn{1}{c}{\%$\Delta$} \\
\midrule
en & ru &  0.348 & 0.249 & ($-$28.4) & 0.366 & ($+$5.2) \\ 
en & de &  0.651 & 0.723 & ($+$11.1) & 0.639 & ($-$1.8) \\ 

en & et &  0.535 & 0.478 & ($-$10.7) & 0.545 & ($+$1.9) \\

en & tr &  0.269 & 0.116 & ($-$56.9) & 0.294 & ($+$9.3) \\ 
en & fi &  0.568 & 0.708 & ($+$24.6) & 0.541 & ($-$4.8) \\

en & cs &  0.616 & 0.612 & ($-$0.6) & 0.617 & ($+$0.2) \\ 
en & zh &  0.241 & 0.244 & ($+$1.2) & 0.241 & ($+$0.0) \\ 
\bottomrule
\end{tabular}

\caption{Evaluation of the best WMT18 and WMT19 submissions (same format as Table~\ref{t:best_all}) using only the sentences that fall within a standard deviation of the mean sentence length $[\mu\pm\sigma]$ for each dataset.
}
\label{t:eval_bucket_1}
\vspace{-1em}
\end{table}

\begin{table}[ht!]
\small
\centering
\begin{tabular}{r@{$\rightarrow$}l rr@{\ }rr@{\ }r}
\toprule
\multicolumn{7}{c}{Translation \textit{into} English}\\
\multicolumn{2}{c}{}
  & \multicolumn{1}{c}{all} &
    \multicolumn{2}{c}{w/ neg.} &
    \multicolumn{2}{c}{w/o neg.} \\ \cmidrule(lr){3-3} \cmidrule(lr){4-5} \cmidrule(lr){6-7}
\multicolumn{2}{c}{WMT18} &
  \multicolumn{1}{c}{Z} & \multicolumn{1}{c}{Z} & \multicolumn{1}{c}{\%$\Delta$} & \multicolumn{1}{c}{Z} & \multicolumn{1}{c}{\%$\Delta$} \\
\midrule

ru & en &  0.25 & 0.65 & ($+$160.0) & 0.21 & ($-$16.0)\\
de & en &  0.49 & 0.31 & ($-$36.7) & 0.51 & ($+$4.1) \\ 
et & en &  0.34 & 0.27 & ($-$20.6) & 0.35 & ($+$2.9)\\ 

tr & en &  0.12 & 0.33 & ($+$175.0) & 0.09 & ($-$25.0)\\
fi & en &  0.18 & 0.53 & ($+$194.4) & 0.13 & ($-$27.7)\\

cs & en &  0.27 & 0.20 & ($-$25.9) & 0.28 & ($+$3.7)\\ 
zh & en &  0.08 & 0.44 & ($+$450.0) & 0.06 & ($-$25.0)\\  
\midrule
\multicolumn{2}{c}{WMT19} \\
ru & en & 0.26 & 0.09 & ($-$65.3) & 0.28 & ($+$7.7) \\
de & en & 0.11 & -0.02 & ($-$118.2) & 0.12 & ($+$9.09) \\

fi & en & 0.33 & 0.61 & ($+$84.8) & 0.31 & ($-$6.1) \\
lt & en & 0.48 & 0.40 & ($-$16.7) & 0.48 & ($+$0.0) \\

gu & en & 0.21 & -0.41 & ($-$295.2) & 0.22 & ($+$4.8) \\
kk & en & 0.25 & 0.05 & ($-$80.0) & 0.27 & ($+$8.0) \\

\end{tabular}
\begin{tabular}{r@{$\rightarrow$}l rr@{\ }rr@{\ }r}
\toprule
\multicolumn{7}{c}{Translation \textit{from} English}\\ 
\multicolumn{2}{c}{}
  & \multicolumn{1}{c}{all} &
    \multicolumn{2}{c}{w/ neg.} &
    \multicolumn{2}{c}{w/o neg.} \\ 
    \cmidrule(lr){3-3} \cmidrule(lr){4-5} \cmidrule(lr){6-7} 
\multicolumn{2}{c}{WMT18} &
  \multicolumn{1}{c}{Z} & \multicolumn{1}{c}{Z} & \multicolumn{1}{c}{\%$\Delta$} & \multicolumn{1}{c}{Z} & \multicolumn{1}{c}{\%$\Delta$} \\
\midrule
en & ru &  0.44 & 0.42 & ($-$4.5) & 0.44 & ($+$0.0) \\ 
en & de &  0.71 & 0.36 & ($-$49.3) & 0.76 & ($+$7.0) \\ 

en & et &  0.76 & 0.53 & ($-$30.3) & 0.78 & ($+$2.6) \\
en & tr &  0.66 & 0.57 & ($-$13.6) & 0.68 & ($+$3.0) \\ 

en & fi &  0.54 & 0.50 & ($-$7.4) & 0.55 & ($+$1.9) \\ 
en & cs &  0.56 & 0.46 & ($-$17.9) & 0.58 & ($+$3.6) \\ 

en & zh &  0.01 & 0.25 & ($+$2400.0) & 0.00 & ($-$100.0) \\ 
\bottomrule
\end{tabular}

\caption{Evaluation of the best WMT18 and WMT19 submissions (same format as Table~\ref{t:best_all}) using only the sentences shorter than one standard deviation of the average $(0,\mu-\sigma)$ for each dataset.
}
\label{t:eval_bucket_3}
\vspace{-1em}
\end{table}

\begin{table}[ht!]
\small
\centering
\begin{tabular}{r@{$\rightarrow$}l rr@{\ }rr@{\ }r}
\toprule
\multicolumn{7}{c}{Translation \textit{into} English}\\
\multicolumn{2}{c}{}
  & \multicolumn{1}{c}{all} &
    \multicolumn{2}{c}{w/ neg.} &
    \multicolumn{2}{c}{w/o neg.} \\ \cmidrule(lr){3-3} \cmidrule(lr){4-5} \cmidrule(lr){6-7}
\multicolumn{2}{c}{WMT18} &
  \multicolumn{1}{c}{Z} & \multicolumn{1}{c}{Z} & \multicolumn{1}{c}{\%$\Delta$} & \multicolumn{1}{c}{Z} & \multicolumn{1}{c}{\%$\Delta$} \\
\midrule

ru & en &  0.08 & 0.03 & ($-$62.5) & 0.09 & ($+$12.5)\\
de & en &  0.40 & 0.29 & ($-$27.5) & 0.44 & ($+$10.0) \\ 
et & en &  0.31 & 0.20 & ($-$35.5) & 0.34 & ($+$9.6)\\ 
tr & en &  -0.14 & -0.18 & ($+$28.6) & -0.13 & ($-$7.1)\\
fi & en &  0.15 & 0.29 & ($+$93.3) & 0.12 & ($-$20.0\\
cs & en &  0.31 & 0.26 & ($-$16.1) & 0.32 & ($+$3.2)\\ 
zh & en &  0.05 & 0.04 & ($-$20.0) & 0.05 & ($+$0.0)\\  
\midrule
\multicolumn{2}{c}{WMT19} \\
ru & en & 0.15 & 0.21 & ($+$40.0) & 0.13 & ($-$13.3) \\
de & en & -0.07 & -0.05 & ($-$28.6) & -0.07 & ($+$0.0) \\
fi & en & 0.21 & 0.38 & ($+$81.0) & 0.18 & ($-$14.3) \\
lt & en & 0.14 & 0.13 & ($-$7.1) & 0.14 & ($+$0.0) \\
gu & en & 0.18 & 0.15 & ($-$16.7) & 0.18 & ($+$0.0) \\
kk & en & 0.16 & -0.13 & ($-$181.3) & 0.18 & ($+$12.5) \\

\end{tabular}
\begin{tabular}{r@{$\rightarrow$}l rr@{\ }rr@{\ }r}
\toprule
\multicolumn{7}{c}{Translation \textit{from} English}\\ 
\multicolumn{2}{c}{}
  & \multicolumn{1}{c}{all} &
    \multicolumn{2}{c}{w/ neg.} &
    \multicolumn{2}{c}{w/o neg.} \\ 
    \cmidrule(lr){3-3} \cmidrule(lr){4-5} \cmidrule(lr){6-7} 
\multicolumn{2}{c}{WMT18} &
  \multicolumn{1}{c}{Z} & \multicolumn{1}{c}{Z} & \multicolumn{1}{c}{\%$\Delta$} & \multicolumn{1}{c}{Z} & \multicolumn{1}{c}{\%$\Delta$} \\
\midrule
en & ru &  0.29 & 0.14 & ($-$51.7) & 0.33 & ($+$13.8) \\ 
en & de &  0.62 & 0.72 & ($+$16.1) & 0.57 & ($-$8.1) \\
en & et &  0.42 & 0.28 & ($-$33.3) & 0.47 & ($+$11.9) \\
en & tr &  -0.06 & -0.34 & ($+$466.7) & 0.00 & ($-$100.0) \\
en & fi &  0.26 & 0.02 & ($-$92.3) & 0.32 & ($+$23.1) \\ 
en & cs &  0.52 & 0.50 & ($-$3.8) & 0.53 & ($+$1.9) \\ 
en & zh &  0.28 & 0.17 & ($-$39.3) & 0.31 & ($+$10.7) \\ 
\bottomrule
\end{tabular}

\caption{Evaluation of the best WMT18 and WMT19 submissions (same format as Table~\ref{t:best_all}) using only the  sentences longer than one standard deviation of the average $(\mu+\sigma,+\infty)$ for each dataset.
}
\label{t:eval_bucket_2}
\vspace{-1em}
\end{table}
\section{Z-score vs. Metrics: WMT19}
\label{s:metrics_app}

\begin{table*}[ht!]
\small
\centering
\begin{tabular}{r@{ $\rightarrow$ }l rr rr rr}
\toprule

\multicolumn{2}{c}{} & \multicolumn{2}{c}{Z vs. BLEU} & \multicolumn{2}{c}{Z vs. chrF++} & \multicolumn{2}{c}{Z vs. METEOR} \\
  \cmidrule(lr){3-4} \cmidrule(lr){5-6} \cmidrule(lr){7-8}
\multicolumn{2}{c}{} & w/ neg. & w/o neg. & w/ neg. & w/o neg. & w/ neg. & w/o neg. \\

\midrule

ru & en & 0.420 & 0.714 & 0.508 & 0.736 & 0.464 & 0.604 \\ \midrule
de & en & 0.483 & 0.633 & 0.517 & 0.600 & 0.517 & 0.567 \\ \midrule
fi & en & 0.788 & 0.818 & 0.848 & 0.909 & 0.879 & 0.758 \\ \midrule
lt & en & 0.709 & 0.796 & 0.673 & 0.796 & 0.709 & 0.796 \\ \midrule
gu & en & 0.564 & 0.745 & 0.782 & 0.964 & 0.527 & 0.891 \\ \midrule
kk & en & 0.855 & 0.818 & 0.673 & 0.855 & 0.745 & 0.782 \\
\bottomrule
\end{tabular}
\caption{
  Comparison of WMT19 rankings (Kendall's $\tau$) using several metrics, and sentences with and without negation.
  We compare the official scoring (normalized direct assessments, Z-score)
  and three automated metrics: BLEU, chrF++ and METEOR.
  All $\tau$ coefficients are statistically significant ($p<0.05$)
  (null hypothesis: there is no association between the Z-score and the automatic metric).
  The differences in the rankings obtained with the Z-score and the three metrics are more substantial when negation is present (lower $\tau$ coefficients).
  }
\label{t:da_bleu_meteor}
\vspace{-1em}
\end{table*}

In Section~\ref{sec:quant}, we discuss correlations between Z-scores and three widely-used automated metrics for assessing the quality of machine translation outputs using the data from WMT18.
Table \ref{t:da_bleu_meteor} shows the same analysis using the data from WMT19.
The conclusions are the same.

\section{Ranking of Submissions}
\label{s:ranking_of_systems}

In Section~\ref{sec:quant}, we show the ranking of systems obtained with all sentences, sentences with negation, and sentences without negation translating from Russian to English.
In this section, we show the rankings of a few other language directions (Estonian to English in Table \ref{t:ranking_et_en}, Chinese to English in Table \ref{t:ranking_zh_en}, Lithuanian to English in Table \ref{t:ranking_lt_en}, Gujarati to English in Table \ref{t:ranking_gu_en} and English to Turkish in Table \ref{t:ranking_en_tr}).
We observe again many changes in the rankings calculated with sentences containing negation.

\begin{table*}[]
\small
\centering
\begin{tabular}{l@{\ }rlr
                r clr
                r clr}
\toprule

& 
\multicolumn{3}{c}{All sentences} &
\multicolumn{4}{c}{Sentences w/ negation} &
\multicolumn{4}{c}{Sentences w/o negation} \\ \cmidrule(lr){2-4} \cmidrule(lr){5-8} \cmidrule(lr){9-12}
& & \multicolumn{1}{c}{System} & \multicolumn{1}{c}{Z} &
  & & \multicolumn{1}{c}{System} & \multicolumn{1}{c}{Z} &
  & & \multicolumn{1}{c}{System} & \multicolumn{1}{c}{Z} \\

\midrule

\multirow{8}{*}{\begin{sideways}WMT18\end{sideways}}
& 1 & tilde-nc-nmt & 0.326  
& 1 & -- & tilde-nc-nmt & 0.232 
& 1 & -- & tilde-nc-nmt & 0.343 \\ 

& 2 & NICT & 0.238 
& 2 &  \up{3} & uedin & 0.179
& 2 & -- & NICT & 0.255 \\ 

& 3 & tilde-c-nmt & 0.215 
& 3 & -- & tilde-c-nmt & 0.167 
& 3 & -- & tilde-c-nmt & 0.224 \\ 

& 4 & M4t1ss & 0.187
& 4 & \down{2} & NICT & 0.146
& 4 & -- & M4t1ss & 0.207 \\ 

& 5 & uedin & 0.186 
& 5 & \up{1} & tilde-comb & 0.129
& 5 & -- & uedin & 0.187 \\ 

& 6 & tilde-comb & 0.171
& 6 & \up{4} & online-A & 0.116
& 6 & -- & tilde-comb & 0.179 \\ 

& 7 & online-B & 0.117 
& 7 & \down{3} & M4t1ss & 0.075 
& 7 & \up{2} & talp-upc & 0.134 \\ 

& 8 & HY-NMT-et-en & 0.106 
& 8 & -- & HY-NMT-et-en & 0.065 
& 8 & \down{1} & online-B & 0.13 \\

& 9 & talp-upc & 0.106 
& 9 & \up{2} & CUNI-Kocmi & 0.061 
& 9 & \down{1} & HY-NMT-et-en & 0.114 \\

& 10 & online-A & 0.063 
& 10 & \down{3} & online-B & 0.043
& 10 & -- & online-A & 0.053 \\

& 11 & CUNI-Kocmi & 0.007
& 11 & \down{2} & talp-upc & -0.039
& 11 & -- & CUNI-Kocmi & -0.003 \\

& 12 & neurotolge.ee & -0.117
& 12 & -- & neurotolge.ee & -0.126
& 12 & -- & neurotolge.ee & -0.116 \\

& 13 & online-G & -0.341
& 13 & -- & online-G & -0.351
& 13 & -- & online-G & -0.339 \\

& 14 & UnsupTartu & -0.950
& 14 & -- & UnsupTartu & -0.964
& 14 & -- & UnsupTartu & -0.948 \\

\bottomrule

\end{tabular}
\caption{Rankings of all submissions translating from Estonian to English using
all sentences (official WMT 2018 ranking),
sentences with negation,
and sentences without negation. tilde-comb refers to tilde-c-nmt-comb.
}
\label{t:ranking_et_en}
\end{table*}

\begin{table*}[]
\small
\centering
\begin{tabular}{l@{\ }rlr
                r clr
                r clr}
\toprule

& 
\multicolumn{3}{c}{All sentences} &
\multicolumn{4}{c}{Sentences w/ negation} &
\multicolumn{4}{c}{Sentences w/o negation} \\ \cmidrule(lr){2-4} \cmidrule(lr){5-8} \cmidrule(lr){9-12}
& & \multicolumn{1}{c}{System} & \multicolumn{1}{c}{Z} &
  & & \multicolumn{1}{c}{System} & \multicolumn{1}{c}{Z} &
  & & \multicolumn{1}{c}{System} & \multicolumn{1}{c}{Z} \\

\midrule

\multirow{8}{*}{\begin{sideways}WMT18\end{sideways}}
& 1  &  NiuTrans & 0.14 & 1 & \up{8} & UMD & 0.197 & 1 & -- & NiuTrans & 0.137  \\
& 2  &  online-B & 0.111 & 2 & -- & online-B.0 & 0.18 & 2 & \up{2} & Unisound & 0.108  \\
& 3  &  UCAM & 0.109 & 3 & \up{4} & Li-Muze & 0.179 & 3 & -- & UCAM & 0.104  \\
& 4  &  Unisound & 0.108 & 4 & \up{2} & Unisound & 0.17 & 4 & \down{2} & online-B & 0.103  \\
& 5  &  Tencent-ens & 0.099 & 5 & \down{4} & NiuTrans & 0.162 & 5 & -- & Tencent-ens & 0.092  \\
& 6  &  Unisound & 0.094 & 6 & \down{1} & Tencent-ens & 0.156 & 6 & -- & Unisound & 0.083  \\
& 7  &  Li-Muze & 0.091 & 7 & \down{4} & UCAM & 0.149 & 7 & \up{1} & NICT & 0.083  \\
& 8  &  NICT & 0.089 & 8 & -- & NICT & 0.13 & 8 & \down{1} & Li-Muze & 0.08  \\
& 9  &  UMD & 0.078 & 9 & \down{5} & Unisound & 0.105 & 9 & -- & UMD & 0.063  \\
& 10  &  online-Y & -0.005 & 10 & -- & online-Y & 0.027 & 10 & -- & online-Y & -0.01  \\
& 11  &  uedin & -0.017 & 11 & \up{1} & online-A & -0.003 & 11 & -- & uedin & -0.018  \\
& 12  &  online-A & -0.061 & 12 & \down{1} & uedin & -0.012 & 12 & -- & online-A & -0.069  \\
& 13  &  online-G & -0.327 & 13 & \up{1} & online-F & -0.329 & 13 & -- & online-G & -0.323  \\
& 14  &  online-F & -0.377 & 14 & \down{1} & online-G & -0.354 & 14 & -- & online-F & -0.383  \\

\bottomrule

\end{tabular}

\caption{Rankings of all submissions translating from Chinese to English using
all sentences (official WMT 2018 ranking),
sentences with negation,
and sentences without negation. Tencent-ens refers to the Tencent-ensemble-system.
}
\label{t:ranking_zh_en}
\end{table*}

\begin{table*}[]
\small
\centering
\begin{tabular}{l@{\ }rlr
                r clr
                r clr}
\toprule

& 
\multicolumn{3}{c}{All sentences} &
\multicolumn{4}{c}{Sentences w/ negation} &
\multicolumn{4}{c}{Sentences w/o negation} \\ \cmidrule(lr){2-4} \cmidrule(lr){5-8} \cmidrule(lr){9-12}
& & \multicolumn{1}{c}{System} & \multicolumn{1}{c}{Z} &
  & & \multicolumn{1}{c}{System} & \multicolumn{1}{c}{Z} &
  & & \multicolumn{1}{c}{System} & \multicolumn{1}{c}{Z} \\

\midrule

\multirow{8}{*}{\begin{sideways}WMT19\end{sideways}}
& 1  &  GTCOM & 0.234 & 1 & \up{1} & tilde-nc & 0.186 & 1 & -- & GTCOM & 0.262  \\
& 2  &  tilde-nc & 0.216 & 2 & \up{1} & NEU & 0.147 & 2 & \up{1} & NEU & 0.226  \\
& 3  &  NEU & 0.213 & 3 & \up{1} & MSRA.MASS & 0.134 & 3 & \up{2} & tilde-c-nmt & 0.226  \\
& 4  &  MSRA.MASS & 0.206 & 4 & \down{3} & GTCOM & 0.093 & 4 & \down{2} & tilde-nc & 0.222  \\
& 5  &  tilde-c-nmt & 0.202 & 5 & -- & tilde-c-nmt & 0.084 & 5 & \down{1} & MSRA.MASS & 0.22  \\
& 6  &  online-B & 0.107 & 6 & -- & online-B & 0.013 & 6 & -- & online-B & 0.126  \\
& 7  &  online-A & -0.056 & 7 & -- & online-A & -0.126 & 7 & -- & online-A & -0.043  \\
& 8  &  TartuNLP-c & -0.059 & 8 & -- & TartuNLP-c & -0.141 & 8 & -- & TartuNLP-c & -0.043  \\
& 9  &  online-G & -0.284 & 9 & \up{1} & JUMT & -0.37 & 9 & -- & online-G & -0.261  \\
& 10  &  JUMT.6616 & -0.337 & 10 & \down{1} & online-G & -0.402 & 10 & -- & JUMT & -0.33  \\
& 11  &  online-X & -0.396 & 11 & -- & online-X & -0.531 & 11 & -- & online-X & -0.369  \\

\bottomrule

\end{tabular}

\caption{Rankings of all submissions translating from Lithuanian to English using
all sentences (official WMT 2019 ranking),
sentences with negation,
and sentences without negation. GTCOM and tilde-nc refer to the systems GTCOM-Primary and tilde-nc-nmt, respectively.
}
\label{t:ranking_lt_en}
\end{table*}

\begin{table*}[]
\small
\centering
\begin{tabular}{l@{\ }rlr
                r clr
                r clr}
\toprule

& 
\multicolumn{3}{c}{All sentences} &
\multicolumn{4}{c}{Sentences w/ negation} &
\multicolumn{4}{c}{Sentences w/o negation} \\ \cmidrule(lr){2-4} \cmidrule(lr){5-8} \cmidrule(lr){9-12}
& & \multicolumn{1}{c}{System} & \multicolumn{1}{c}{Z} &
  & & \multicolumn{1}{c}{System} & \multicolumn{1}{c}{Z} &
  & & \multicolumn{1}{c}{System} & \multicolumn{1}{c}{Z} \\

\midrule

\multirow{8}{*}{\begin{sideways}WMT19\end{sideways}}
& 1  &  NEU & 0.210 & 1 & -- & NEU & 0.112 & 1 & -- & NEU & 0.221  \\
& 2  &  UEDIN & 0.126 & 2 & \up{1} & GTCOM & 0.069 & 2 & -- & UEDIN & 0.141  \\
& 3  &  GTCOM & 0.100 & 3 & \up{3} & NICT & 0.037 & 3 & -- & GTCOM & 0.104  \\
& 4  &  CUNI-T2T & 0.090 & 4 & -- & CUNI-T2T & 0.018 & 4 & -- & CUNI-T2T & 0.098  \\
& 5  &  aylien-mt & 0.066 & 5 & \down{3} & UEDIN & -0.014 & 5 & -- & aylien-mt & 0.086  \\
& 6  &  NICT & 0.044 & 6 & \up{2} & IITP-MT & -0.018 & 6 & -- & NICT & 0.045  \\
& 7  &  online-G & -0.189 & 7 & \down{2} & aylien-mt & -0.128 & 7 & -- & online-G & -0.189  \\
& 8  &  IITP-MT & -0.192 & 8 & \down{1} & online-G & -0.192 & 8 & -- & IITP-MT & -0.211  \\
& 9  &  UdS-DFKI & -0.277 & 9 & -- & UdS-DFKI & -0.277 & 9 & -- & UdS-DFKI & -0.277  \\
& 10  &  IIITH-MT & -0.296 & 10 & -- & IIITH-MT & -0.355 & 10 & -- & IIITH-MT & -0.29  \\
& 11  &  Ju-Saarland & -0.598 & 11 & -- & Ju-Saarland & -0.566 & 11 & -- & Ju-Saarland & -0.601  \\

\bottomrule

\end{tabular}

\caption{Rankings of all submissions translating from Gujarati to English using
all sentences (official WMT 2019 ranking),
sentences with negation,
and sentences without negation. CUNI-T2T, aylien-mt, and GTCOM refer to the systems CUNI-T2T-transfer-guen, aylien-mt-gu-en-multilingual, and GTCOM-Primary, respectively
}
\label{t:ranking_gu_en}
\end{table*}

\begin{table*}[]
\small
\centering
\begin{tabular}{l@{\ }rlr
                r clr
                r clr}
\toprule

& 
\multicolumn{3}{c}{All sentences} &
\multicolumn{4}{c}{Sentences w/ negation} &
\multicolumn{4}{c}{Sentences w/o negation} \\ \cmidrule(lr){2-4} \cmidrule(lr){5-8} \cmidrule(lr){9-12}
& & \multicolumn{1}{c}{System} & \multicolumn{1}{c}{Z} &
  & & \multicolumn{1}{c}{System} & \multicolumn{1}{c}{Z} &
  & & \multicolumn{1}{c}{System} & \multicolumn{1}{c}{Z} \\

\midrule

\multirow{8}{*}{\begin{sideways}WMT18\end{sideways}}
& 1  &  online-B & 0.277 & 1 & \up{1} & uedin & 0.231 & 1 & -- & online-B & 0.308  \\
& 2  &  uedin & 0.222 & 2 & \up{1} & alibaba5732 & 0.138 & 2 & \up{1} & alibaba5732 & 0.232  \\
& 3  &  alibaba5732 & 0.216 & 3 & \up{2} & alibaba5744 & 0.127 & 3 & \down{1} & uedin & 0.221  \\
& 4  &  NICT & 0.128 & 4 & -- & NICT & 0.1 & 4 & -- & NICT & 0.135  \\
& 5  &  alibaba5744 & 0.111 & 5 & \down{4} & online-B & 0.094 & 5 & -- & alibaba5744 & 0.107  \\
& 6  &  online-G & 0.058 & 6 & -- & online-G & -0.004 & 6 & -- & online-G & 0.071  \\
& 7  &  RWTH & -0.06 & 7 & -- & RWTH & -0.097 & 7 & -- & RWTH & -0.052  \\
& 8  &  online-A & -0.254 & 8 & -- & online-A & -0.481 & 8 & -- & online-A & -0.206  \\

\bottomrule

\end{tabular}

\caption{Rankings of all submissions translating from English to Turkish using
all sentences (official WMT 2018 ranking),
sentences with negation,
and sentences without negation. alibaba5732 and alibaba5744 refer to the systems alibaba-ensemble-model.5732 and alibaba-ensemble-model.5744, respectively.
}
\label{t:ranking_en_tr}
\end{table*}

\end{document}